\documentclass[letterpaper, 10 pt, conference]{ieeeconf}

\IEEEoverridecommandlockouts                              

\usepackage{graphicx} 
\graphicspath{{figures/}}
\usepackage{optidef} 
\usepackage[sort,compress]{cite}
\usepackage{amsmath} 
\usepackage{amssymb}  
\usepackage{color}
\usepackage[ruled]{algorithm2e}
\usepackage[noend]{algorithmic}
\usepackage{tikz}
\usetikzlibrary{positioning}

\def\vd{8pt}
\def\lessverticaldistance{2pt}

\title{\LARGE \bf
Adaptive Robot Assistance: Expertise and Influence \\in Multi-User Task Planning
}

\author{Abhinav Dahiya and 
Stephen L.\ Smith
\thanks{This research is partially supported by the Natural Sciences and Engineering Research Council of Canada (NSERC). } 
\thanks{The authors are with Department of Electrical and Computer Engineering, University of Waterloo {\tt\small a4dahiya@uwaterloo.ca}, {\tt\small stephen.smith@uwaterloo.ca}}%
}

\begin{document}

\maketitle
\thispagestyle{empty}
\pagestyle{empty}

\begin{abstract}

This paper addresses the challenge of enabling a single robot to effectively assist multiple humans in decision-making for task planning domains. We introduce a comprehensive framework designed to enhance overall team performance by considering both human expertise in making the optimal decisions and robot influence on human decision-making. Our model integrates these factors seamlessly within the task-planning domain, formulating the problem as a partially observable Markov decision process (POMDP) while treating expertise and influence as unobservable components of the system state. To solve for the robot's actions in such systems, we propose an efficient Attention-Switching policy. This policy capitalizes on the inherent structure of such systems, solving multiple smaller POMDPs to generate heuristics for prioritizing interactions with different human teammates, thereby reducing the state space and improving scalability. Our empirical results on a simulated kit fulfillment task demonstrate improved team performance when the robot's policy accounts for both expertise and influence. This research represents a significant step forward in the field of adaptive robot assistance, paving the way for integration into cost-effective small and mid-scale industries, where substantial investments in robotic infrastructure may not be economically viable.
\end{abstract}

\section{Introduction}
Advances in human-robot interaction (HRI) research have yielded highly efficient multi-robot systems that require minimal human intervention. These systems empower a single human operator to coordinate and manage multiple robots concurrently \cite{glas2011teleoperation, goodrich2007managing, dahiya2023survey}. While these systems excel in applications such as industrial automation and service robotics, there exist contexts, including mid/low-scale industries and patient care settings, where practicality and financial constraints limit the feasibility of multi-robot systems. In such scenarios, human workforce remains a pragmatic and cost-effective choice over complete automation. Consequently, our focus shifts from a single human operator managing multiple robots to the more favorable scenario of a single robot assisting multiple humans. This paper addresses the associated challenges and opportunities in task-planning domains.

In this paper, we introduce a multi-human single-robot interaction problem, where a single robot is tasked with providing assistance to multiple human teammates.  The assistance is provided through a Decision Support System (DSS) that conveys the next actions for humans to take. Our framework encompasses two pivotal aspects of human behavior within the human-robot system.  The first aspect models human decision-making policies, enabling the prediction of their future actions. This modeling takes into account various factors, including human expertise \cite{milliken2017modeling,enayati2018skill}, situational awareness \cite{endsley2000theoretical}, workload \cite{prewett2010managing}, and preferences \cite{cakmak2011human}. Understanding these factors empowers the robot to anticipate how humans will act as decision-making agents within a given environment. The second aspect centers on the \textit{influenceability} of humans and how they adapt their behavior in response to the robot's actions and suggestions. This influence is governed by factors such as user trust in the robot \cite{mittu2016robust, sadrfaridpour2016modeling}, cooperation \cite{wu2016trust}, adaptability \cite{nikolaidis2017mathematical}, and comprehension of the robot's actions \cite{dragan2015effects}. These elements significantly shape how human behavior changes based on the actions and suggestions provided by the robot.

\begin{figure}
    \centering
    \includegraphics[width=0.99\columnwidth]{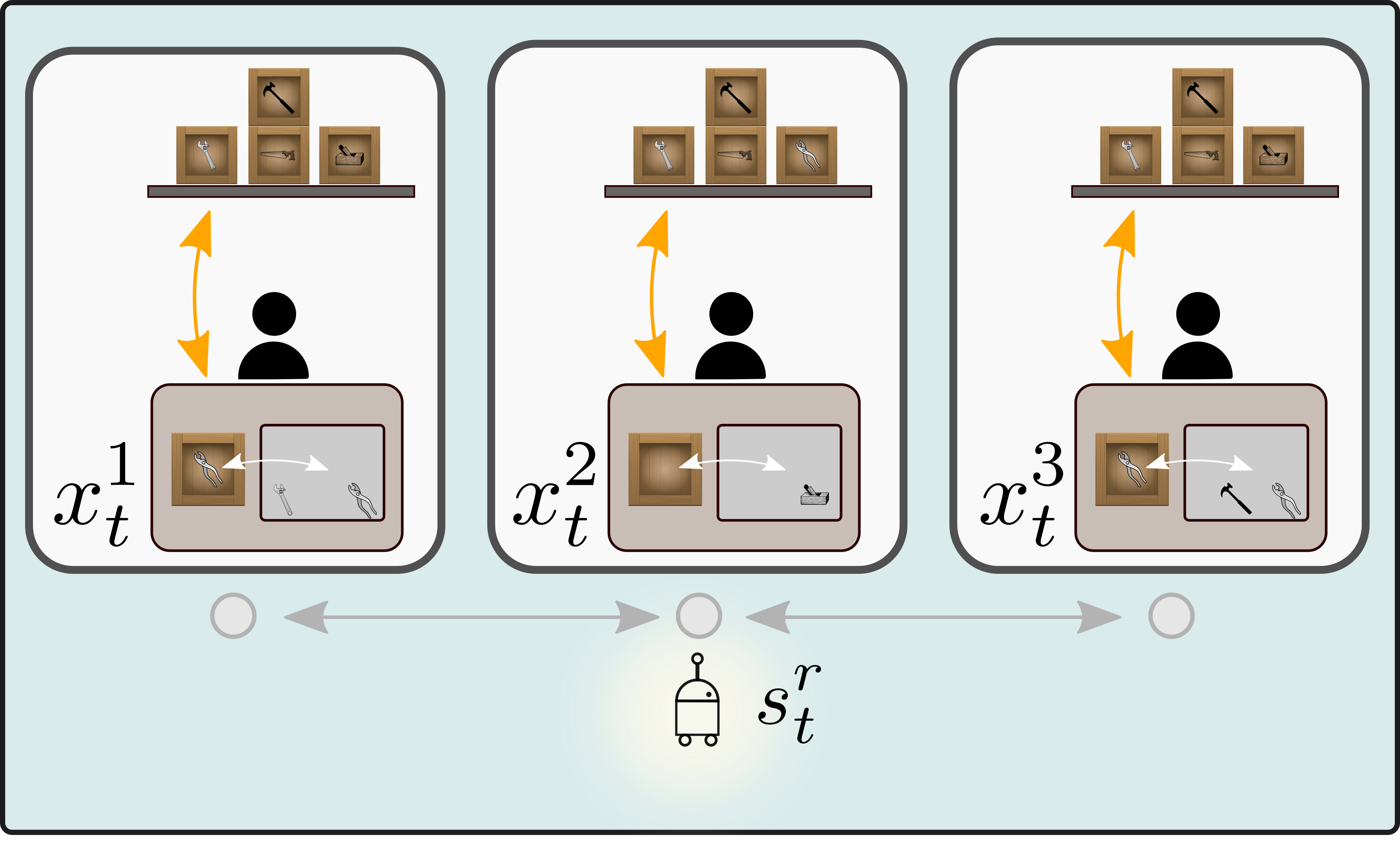}
    \caption{Visualization of the Multi-Human Assistance Problem using a Kit Fulfillment Task. In this scenario, several human workers work independently within their individual environments ($x^i_t$), each aiming to assemble a specified kit in the fewest possible steps. The robot, denoted by its state $s^r_t$, has the ability to move and provide assistance to the human workers through action execution or suggestions to enhance task efficiency. The robot's decision on which human to assist is governed by factors including human expertise and influenceability.} 
    \label{fig:task_setup}
\end{figure}

This paper makes three key contributions: First, we introduce a comprehensive mathematical model for the multi-human single-robot system that seamlessly integrates human interactions with both their environment and the robot. We provide stochastic models that estimate human behavior in task-planning domains, incorporating aspects of expertise and influence.  Second, we propose an efficient Attention-Switching policy designed to address the multi-human interaction problem. This policy divides the problem into independent single-human instances and leverages the problem structure to select which human teammate to assist. Third, we present a simulated kit fulfillment task (illustrated in Figure~\ref{fig:task_setup}) as a practical demonstration of how the proposed method allow multi-human single-robot teams to significantly enhance the performance without necessitating personal robot assistants for each team member.

The kit fulfillment task, also known as a \textit{kitting} task, represents a typical instance of high-level planning problems. Such tasks frequently occur in industries involved in custom product creation or order fulfillment and have previously served as a basis for human expertise estimation within task-planning domains \cite{pamela2019expertise}. In this scenario, the robot's role is to assist human workers in optimally selecting which container to move and assess in order to expedite the kit fulfillment task. In this capacity, the robot faces the challenge of determining which team member requires its assistance the most, assessing the willingness of human teammates to accept this support, and optimizing its actions to minimize the cost of moving between different human workers' locations.


\section{Background and Related Work}
\label{sec:background}

This section provides a brief background on the role of expertise in human decision-making and how it can be used to predict human behavior. We also discuss how the robot's influence on humans can be modelled and how it affects the human-robot interaction outcome.

\subsection{Human as a decision-making agent}

Human behavior in various environments is often modeled as that of a rational decision-making agent with varying degrees of expertise. Human expertise can be defined as the ability to optimize performance metrics \cite{huegel2009expertise}, the capacity to leverage domain knowledge \cite{charness2008role}, or the proficiency to accomplish task objectives independently \cite{milliken2017modeling}.
In the context of task-planning, expertise is frequently viewed as the ability to estimate the true utilities of states and make choices based on those estimates \cite{ramachandran2007bayesian}. Humans estimate the utilities of their actions based on task requirements and personal preferences~\cite{abbeel2004apprenticeship}.
According to the theory of mind, for given state space $S$ and action space $A$, human actions can be modeled through a Boltzmann policy \cite{ramachandran2007bayesian, rafferty2015inferring} as follows:
\begin{align}
    \pi(s, a) = Pr(a | s) = \frac{1}{Z_s} \exp{(-\beta \, Q(s, a))},
\label{eq:boltzmann}
\end{align}
where $Q(s,a)$ is a real-valued function representing the \textit{utility} or \textit{value} of taking action $a \in A$ in a state $s \in S$, and $Z_s$ is an appropriate normalization constant. 
The parameter $\beta$, known as the \textit{expertise coefficient} or precision parameter, accounts for human sub-optimality, especially when there is minimal difference among utility values. These parameters define expertise and predict human actions given a system state. This policy is particularly useful for modeling human limited knowledge of optimal utilities and their sensitivity to differences in these utilities.

\subsection{Influenceability in human-robot interaction}
Human influenceability, akin to trust, is a crucial factor in determining human-robot interaction outcomes \cite{sadrfaridpour2016modeling, chen2018planning}. While trust dynamics have been modeled in various ways, including as a Bayesian network \cite{xu2015optimo}, a linear update \cite{chen2018planning}, and a time series \cite{wang2014human}, influenceability represents the likelihood of a human to accept or agree with their robot partner's actions and decisions \cite{freedy2007measurement}.
In \cite{chen2018planning}, the authors propose a linear Gaussian system using a scalar influence/trust parameter $\theta$ to capture dynamics for task execution:
\begin{align}
    P(\theta_{t+1} |\theta_t, e_{t+1}) = \mathcal{N}(\alpha_{e_{t+1}}\theta_t + \beta_{e_{t+1}}, \sigma_{e_{t+1}}),
\end{align}
where $\alpha_{e_{t+1}}$, $\beta_{e_{t+1}}$, and $\sigma_{e_{t+1}}$ are system parameters learned from user data. Robot performance $e_{t+1}$ indicates whether the robot succeeded in its task in the previous time step.
In domains such as task-planning, the robot's performance may not be immediately perceivable by the human partner. Therefore, different influence models based on human personal beliefs, rather than actual robot performance, may be required. One such model is presented in Section~\ref{sec:humanModel}.

When assisting multiple humans, accounting for both expertise and influence can enhance team performance by helping the robot allocate its assistance to those team members who will benefit the most. In the next section, we model the dynamics of human-robot interaction in a multi-human scenario and formulate an optimization problem to determine the robot's action policy.

\section{Problem Formulation}
\label{sec:problem_formulation}
To formalize the multi-human single-robot assistance problem, we consider a scenario with $K \geq 1$ human users, each independently working on their respective tasks. Each user $k\in {\cal K} \coloneqq \{1,\ldots, K\}$ is required to complete a sequence of tasks to reach its goal.  
Each user is assumed to make sequential decisions, trying to maximize an arbitrary utility function.  While capable of working independently, human actions may not always be optimal. The robot intervenes through interaction, either by executing actions or suggesting actions to humans, aiming to influence their behavior towards globally optimal actions.

The primary objective is for the robot to assist the humans by monitoring their actions and intervening when it is likely to improve overall performance.  This setup is cast as a discrete-time dynamic system defined by:
\begin{equation}
    s_{t+1} = f(s_t, a^r_t),
\end{equation}
where $s_t$ encompasses the state $s^k_t$ of each human $k\in \mathcal{K}$, along with the robot state $s^r_t$. The robot state $s^r_t$ denotes information such as location of the robot in the environment and the action it is currently executing. Given state $s_t$, the robot selects its next action $a^r_t$. The function $f$ characterizes system dynamics, mapping states and robot actions to subsequent system states. 

%
In this framework, we assume that while humans make decisions based on non-deterministic policies to maximize internal utility functions \cite{swamy2020scaled}, their actual choices are influenced by their capacity to evaluate and compare available options \cite{pamela2019expertise}. The operating state of a human user $k$ at time $t$, denoted as $s^k_t = (x^k_t, h^k_t)$, encompasses the current environment state and the human behavioral state, which governs the behavior of human $k$. The state space for human $k$ is denoted as $\mathcal{S}^k$, the state space for the robot is denoted as $\mathcal{S}^r$, and the collective state space for the system is represented as $\pmb{\mathcal{S}} = \mathcal S^1 \times \cdots \times \mathcal S^K \times \mathcal{S}^r$. Importantly, the evolution of a human's state varies according to the actions they take (with a possible influence from actions of the robot).  We now provide a mathematical model for different components of the system.

\subsection{Model of human users}
\label{sec:humanModel}
Based on the discussion presented in Section~\ref{sec:background}, we contend that the behavioral state of humans in a robot-assisted task-planning setting can be aptly captured through two parameters -  expertise and influence, and can be expressed as:
\begin{equation}
    h^k_t = (\beta^k_t, \theta^k_t).
\end{equation}
The parameter $\beta^k \in [0, \infty)$ is the \textit{expertise coefficient}, signifying a user's proficiency in selecting the optimal action given their environment state.  Assuming humans to be rational agents, their decision-making process can be approximated as independent Markov chains.  These Markov chains dictate actions based on the utility function $Q$ associated with the prevailing environment state $x^k_t$. We employ an expertise-based Boltzmann policy \cite{rafferty2015inferring}, as described by equation \eqref{eq:boltzmann}, to quantify the probabilities of human actions:
\begin{align}   
Pr(a^k_t | x^k_t,\beta^k_t) = \frac{e^{-\beta^k_t ~ Q(x^k_t, a^k_t)}}{Z_{x^k_t}}.
\label{eq:boltzmann2}
\end{align}
This model determines the probability of human $k$ taking action $a^k_t$ in the environment state $x^k_t$, in the absence of robot intervention.

The parameter $\theta^k \in [0,1]$ is the \textit{influence coefficient}, characterizes the influence of the robot on human $k$, signifying the likelihood of the human adopting an action suggested by the robot.  In the context of our problem, this parameter mirrors the human's trust in the robot's capacity to offer correct suggestions.  Notably, higher influence amplifies the likelihood of adopting the robot's suggestion while diminishing the likelihood of selecting alternative actions. To encapsulate the effects of influence on human decision-making, we employ Bayesian estimation as follows: 
\begin{equation}
  Pr(a^k_t | x^k_t,\beta^k_t, a^r_t) = \begin{cases}
               \frac{\theta^k_t}{Z_{x^k_t}}e^{-\beta^k_t ~ Q(x^k_t, a^k_t)}&  \text{if } a^k_t = a^r_t\\[\vd]
               \frac{1-\theta^k_{t}}{Z_{x^k_t}}\,e^{-\beta^k_t ~ Q(x^k_t, a^k_t)} & \text{otherwise},
            \end{cases}
\end{equation}
where $Z_{x^k_t} \coloneqq \sum_{a^k_t} Pr(a^k_t | x^k_t,\beta^k_t, a^r_t)$ is the normalizing constant.
Furthermore, as users observe the outcomes of the robot's suggested actions, their confidence in the reliability of the robot's advice undergoes adaptations. We propose that human assessment of the robot's effectiveness hinges on the degree to which the robot's actions align with their pre-existing beliefs of state utilities.  Similar to existing studies in the literature, where human trust is considered a linear variable and changes in a linear fashion \cite{chen2018planning}, we use the following function to define the change in $\theta$:
\begin{align}
    \theta^k_{t+1} = \theta^k_t + \eta\, \Big(\max_{a^k}\big(Q(x^k_{t+1}, a^k)\big) - \max_{a^k}\big(Q(x^k_{t}, a^k)\big)\Big),
    \label{eq:update_trust}
\end{align}
where $x^k_{t+1}$ is the state that human $k$ is expected to transition to should they follow the robot's suggested action.
This linear update captures the notion that trust tends to improve if the human anticipates an increase in the utility of the next state compared to the current state, and conversely, it diminishes if such improvement is not perceived.

\subsection{System dynamics}
\begin{figure}[t]
    \centering
    \includegraphics[width=0.65\columnwidth]{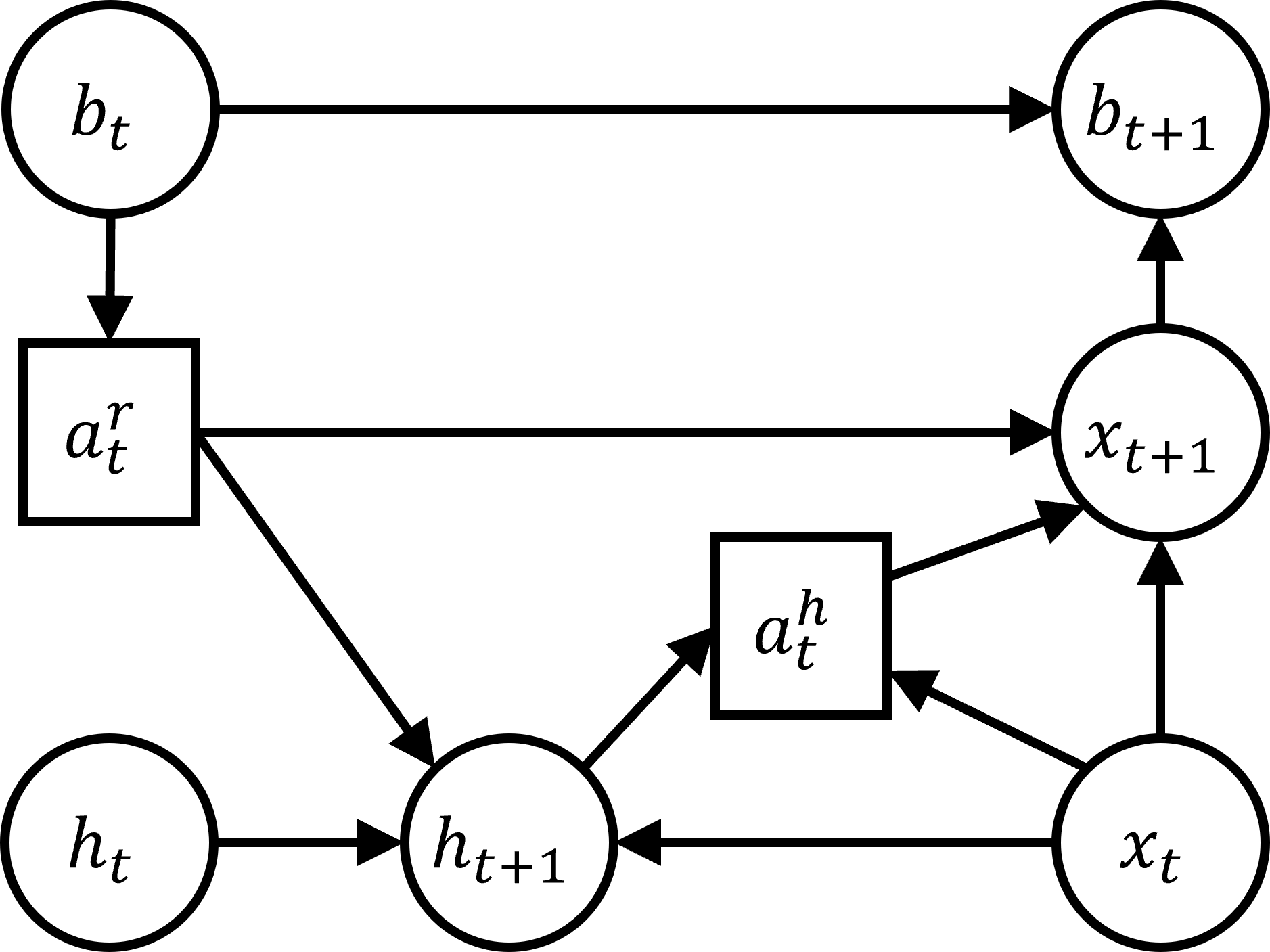}
    \caption{A Dynamic Bayesian Network representation of the presented multi-human assistance POMDP. $x_t$ and $h_t$ respectively represent environment states and behavioral states for all humans. Starting with robot's belief $b_t$, the robot takes an action $a^r_t$. Following that, the human parameters are changed to $h_{t+1}$ as a result of the interaction. Then the human users execute their actions $a^h_t$ to change the environment state to $x_{t+1}$. Robot observes this outcome and updates its belief over the states to $b_{t+1}$.}
    \label{fig:POMDPGraph}
\end{figure}
The control input in our problem is the robot action $a^r_t$ drawn from the action set $\mathcal{A}^r$. These robot actions encompass movements within the environment, leading to changes in the robot state $s^r_t$, as well as providing assistance to human teammates through suggested actions. The robot decides its actions given the system state $s_t$. However, the robot only has complete observability over the environmental state $x_t$, denoted as $x_t=\{x^1_t, \ldots, x^k_t\}$, while human behavioral states remain unobservable and are indirectly estimated through their actions. To guide its decisions, the robot maintains a probability distribution known as the belief $b_t$ over potential system states.

Our problem's system dynamics can be effectively represented using a Dynamic Bayesian Network (DBN), as illustrated in Figure~\ref{fig:POMDPGraph}.  The system reaches its terminal state when each of the $k$ humans complete their respective tasks.

Following each transition, the system receives a reward $R(s_t, a^r_t)$, which is composed of rewards from the robot's own actions, denoted as $R^r(s^r_t, a^r_t)$, and rewards contributed by each human, denoted as $R^k(s^k_t, a^r_t)$. The net system reward is calculated as the sum of these individual rewards:
\begin{equation}
    R(s_t, a^r_t) = R^r(s^r_t, a^r_t) + \sum_{k=1}^K R^k(s^k_t, a^r_t).
    \label{eq:net_reward}
\end{equation}
These reward functions can be designed to align with specific performance criteria, such as minimizing completion time or reducing robot interventions.
The expected total reward earned by a policy $\pi$ is given by
	\begin{equation}
		J(\pi) = \mathbb{E}^\pi \biggl[\sum_{t=0}^{\infty} \gamma^t R(s_t, a^r_t) \biggm| \boldsymbol S_0 = \boldsymbol s_0 \biggr],
		\label{eq:policy-cost}
	\end{equation}
	where $\gamma \in (0,1)$ is the discount factor and $\boldsymbol s_0=(s^1_0,\ldots,s^K_0, s^r_0)$ is the initial state of all users and the robot.

\subsection{Problem Objective}
The design objective is to solve the following optimization problem:

\noindent \textbf{Problem 1.} 
Given the set $\mathcal{K}$ of human users, the system dynamics, the per-step rewards, and the discount factor $\gamma$, choose a policy $\pi : \pmb{\mathcal{S}} \to \mathcal{A}^r$ to maximize the total expected discounted reward $J(\pi)$ given by equation~\eqref{eq:policy-cost}.\vspace{0.55em}

Optimal solution for Problem~1 can be found by modelling it as a Partially Observable Markov Decision Process (POMDP) and solving using dynamic programming~\cite{braziunas2003pomdp}. However, there are several challenges in solving for the exact solution. The state space of the complete system is a product of individual human states, which grows exponentially with the number of human users:
\begin{equation}
    |\pmb{\mathcal{S}}| = \mathcal{O} \big(|\mathcal{S}^1|\,|\mathcal{S}^2|\,\cdots\,|\mathcal{S}^K|\,|\mathcal{S}^r| \big),
\end{equation}
This means the sizes of state and action spaces of the resulting model grows exponentially with the number of humans in the system. Thus, solving Problem~1 using dynamic programming becomes intractable for larger systems as shown in \cite{dahiya2022scalable}. The following section presents a solution technique that addresses this problem by casting it into smaller POMDPs and employing a greedy attention-switching strategy.


\section{Greedy Attention-Switching Policy}
\label{sec:solution}
In this section, we outline our approach, which combines a greedy attention-switching technique with a POMDP solver to address the given problem efficiently, capitalizing on the inherent problem structure. The proposed policy leverages the fact that in the proposed multi-human setting, each human teammate operates as an independent Markov chain, influenced only by direct interactions with the robot.

We initiate our approach by modeling the robot interaction problem with each individual human as separate POMDPs. The robot maintains a belief state $b_t = \{b^1_t, b^2_t, \ldots, b^k_t\}$ for each human's state and makes decisions based on these beliefs. Observations include exact information about the robot state $s^r_t$, the environment state $x_t$, and the actions $a^h_t$ taken by each human.

The proposed policy is outlined in Algorithm~\ref{algo}. This approach avoids the complexity of solving a combined POMDP by initially addressing single human-robot interaction problems for each human and obtaining individual policies $\pi^k$. 

At each time step, the algorithm computes two metrics for each human $k$: 

\noindent 1) The expected discounted reward if the robot executes policy $\pi^k$ for human $k$, denoted as $R^k_\pi$, and 

\noindent 2) The expected discounted reward if the robot does not interact with human $k$, denoted as $R^k_\phi$.

The algorithm then employs these metrics as heuristics to make a greedy selection of the robot's next action, aiming to maximize the expected net reward in the next step (line \ref{line:best_action} of Algorithm \ref{algo}). This decision takes into account not only which human would benefit most from the robot's assistance but also considers the cost of relocating the robot from one human to another. In essence, the algorithm determines whether it is more rewarding to switch the robot's attention to a different human or to continue assisting the current one. After the robot takes its action and observes the outcomes, it updates its belief about the true system state, proceeding to the next time step.
%

\begin{algorithm}
  \begin{algorithmic}[1]
    \STATE Obtain individual policies $\pi^k$ for each human $k$
    \FOR {each human $k$}
      \STATE Initialize belief state $b^k$
    \ENDFOR
    \WHILE{terminal state not reached}
    \FOR{each human $k$}
      \STATE $a^{r_k} \leftarrow \pi^k(b(s^k))$ 
      \STATE $R^k_\phi \leftarrow \sum_{s^k} \,b(s^k)\, R^k(s^k, \phi)$\\[\lessverticaldistance]
      \STATE $R^k_\pi \leftarrow \sum_{s^k} \,b(s^k)\, R^k(s^k, a^{r_k})$\\[\lessverticaldistance]
    \ENDFOR
    \STATE Execute action $a^{r_k} \leftarrow \arg\mathop{\max}_{a^{r_k}} \big[ R^k_\pi - R^k_\phi + R^r(s^r, a^{r_k}) \big]$ \label{line:best_action}
    \FOR {each human $k$}
      \STATE Observe human action $a^k$
      \STATE Update belief $b^k$ based on $s^k, a^{r_k}$ and $a^k$
    \ENDFOR 
    \ENDWHILE
  \end{algorithmic}
  \caption{Greedy Attention-Switching Policy}
  \label{algo}
\end{algorithm}

The key advantage of this approach is that it reduces the problem to solving $K$ smaller POMDPs with a state space exponentially smaller in size compared to the complete system state. This results in solving for the policy in a state space that grows linearly with the number of humans:
\begin{equation}
    |\pmb{\mathcal{S}}| = \mathcal{O} \big( \big(|\mathcal{S}^1|+|\mathcal{S}^2|+\ldots+|\mathcal{S}^K| \big)\, |\mathcal{S}^r| \big),
\end{equation}
In practice, this approach will greatly benefit when several of the human teammates are in a similar environment and working on similar tasks, as the computed offline policies $\pi^i$ can be shared among the teammates.

In the next section, we demonstrate the utility of this policy using the example of a kit fulfillment task.

\section{Kit Fulfillment Task}
\label{sec:results}
The kit fulfillment task is a crucial component of our study, serving as an illustrative example for the multi-human single-robot assistance problem. In this task, $K$ human workers are responsible for assembling specific kits, with the primary objective of minimizing the number of actions required to complete each kit.
\begin{figure}[ht]
\centering
    \includegraphics[width=0.9\columnwidth]{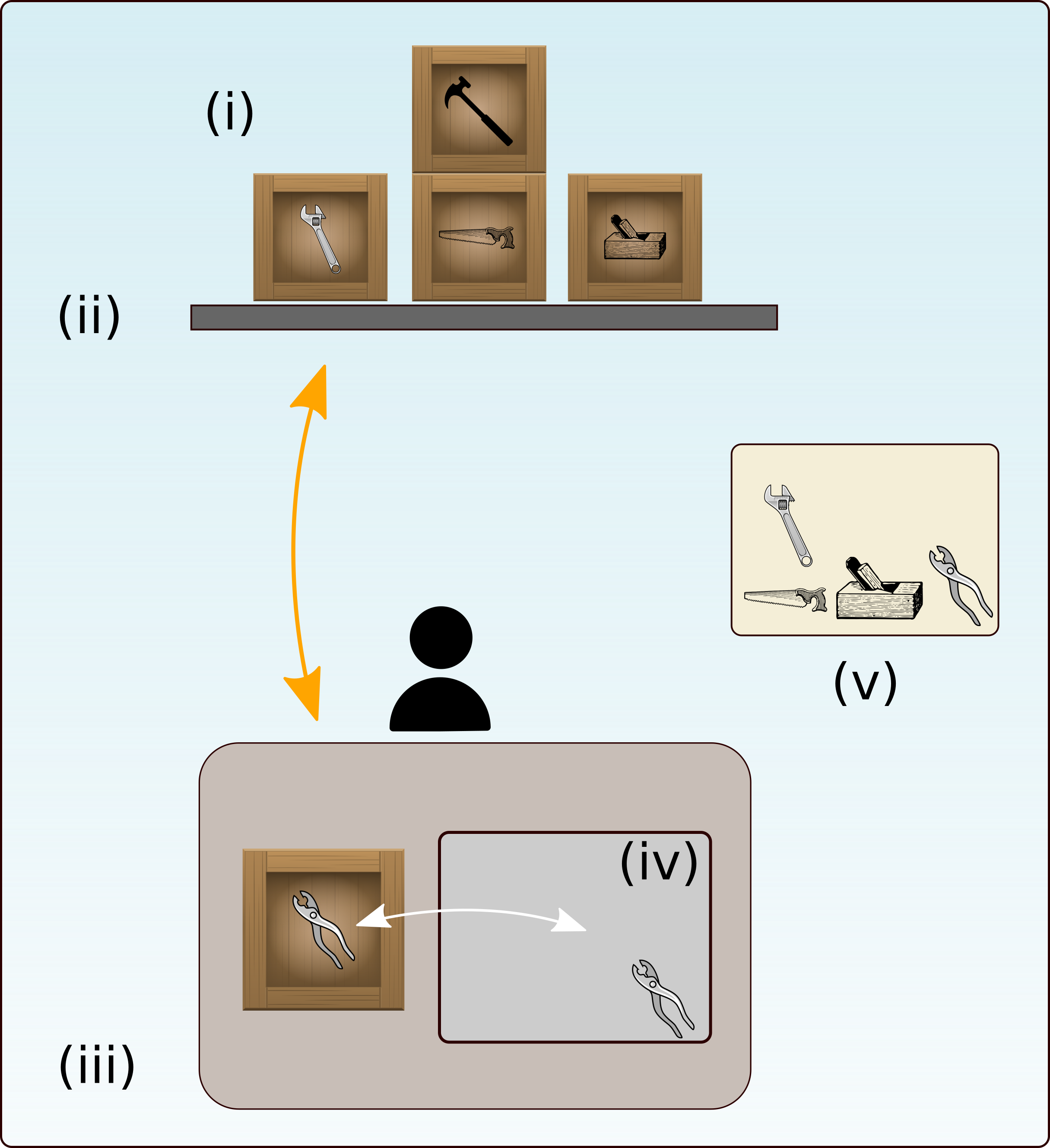}
    \caption{Illustration of the simulated kit fulfillment task for a single human, showing the simulation elements: (i) tool containers, (ii) storage, (iii) unloading platform, (iv) tool-kit box and (v) the target kit arrangement.  The human worker can transport one container at a time between the storage and unloading platform (indicated by the orange arrow) and move a single item per action (indicated by the white arrow). Importantly, a maximum of one container can occupy the unloading platform at any given moment. The task concludes when the tool kit mirrors the target arrangement, and the containers are returned to their initial positions.\vspace{-0.5em}}
\label{fig:kitting_task}
\end{figure}

\subsection{Task Description}
The task involves assembling \textit{kits}, which can include various components, into a predefined arrangement. Each human worker is capable of performing actions such as moving containers and placing items into the kit box. The robot assistant's role is to enhance the efficiency of human workers by suggesting actions, such as relocating containers or adding items to the kits. To make these suggestions, the robot must position itself near the relevant human worker. The task concludes when all human workers have successfully assembled their kits in the required arrangement.

This kit fulfillment task draws inspiration from a previous user study aimed at evaluating human expertise in similar contexts \cite{pamela2019expertise}. This study has highlighted the challenge of accurately assessing the quality of actions, making the task inherently complex. Therefore, it serves as an ideal testing ground for assessing the impact of robot assistance in such task-planning domains.

\subsection{Task Components}
Figure~\ref{fig:kitting_task} provides a visual representation of the kit fulfillment task for a single human, where the goal is to complete the required kit (shown in the inset) in minimum number of actions. Human workers can transport one container at a time or move a single item during each action. Importantly, only one container can be on the unloading platform simultaneously. 
At the end of the task, the kit box must contain the items in the specified arrangement, and all containers must return to their original locations. These specifications can easily be encoded using frameworks like STRIPS or PDDL \cite{aeronautiques1998pddl}.

From a POMDP perspective, the kit fulfillment task represents the observable state ($x$), including the locations of containers, food items, and the robot. The human workers' behavioral parameters make the unobservable part of the system state. To minimize robot intervention while optimizing task completion, costs are associated with robot interventions and robot movements between locations, encoded as a negative value for $R^r(s_t,a^r_ t)$. The problem objective is to minimize the expected sum of the total number of human actions required to complete the task and the number of times the robot needs to move between the workers.

\subsection{Simulations}
To evaluate and validate our proposed solution method for the multi-human single-robot assistance problem, we conducted a series of simulations. In these simulations, we created a diverse pool of human workers, each characterized by different behavioral parameters. This pool represents a range of humans with varying levels of expertise and initial influence of the robot. The utility $Q$ for each worker is generated based on noisy estimates of the \textit{cost-to-go} from a given state to the goal with noise standard deviation $\sigma$. The values chosen for these parameters are summarized in Table~\ref{table_values}. In practice, the human utility function $Q$ and expertise coefficient $\beta$ can be determined by observing human actions \cite{pamela2019expertise}. Similarly, the influence/trust dynamics parameters can be learned through interaction data \cite{chen2018planning, xu2015optimo}.  However, for the purposes of this paper, we treat these parameters as inputs to the problem. This approach allows us to explore a wide range of scenarios by randomly sampling values of the expertise and influence coefficients, reflecting the diversity of human behaviors and perceptions in real-world collaborative environments. The problem and the solution framework was implemented using POMDPs.jl library in Julia \cite{egorov2017pomdps}.

\renewcommand{\arraystretch}{1.5}
\begin{table}[ht]
\centering
\begin{tabular}[pos=c]{|c|c|c|c|c|c|c} 
 \hline
 Parameter & $\beta_{range}$ & $\theta_{range}$ & $\sigma_{range}$ & $\eta$\\ [0.1ex] 
 \hline
 Value & $[0.1,3.0]$ & $[0.5,0.95]$ & $[0.1,1.0]$ & $0.2$\\[0.1ex] 
 \hline
\end{tabular}
\caption{Parameter values used in kit fulfilment task.}
\label{table_values}
\vspace{-1em}
\end{table}
\vspace{-1em}
\subsection{Baseline Policies}
As discussed in Section~\ref{sec:problem_formulation}, the given problem can be theoretically solved by formulating it as a POMDP. However, practical constraints led us to explore alternative approaches, as the POMDP formulation became intractable and failed to find solutions in reasonable amount of time, even for relatively small problem instances during simulations. Therefore, to evaluate the effectiveness of our proposed policy, we compared it against the following two baseline policies: 

\textbf{1) No Robot Assistance (Base Case):} In this baseline scenario, the robot does not provide any assistance to human workers. This scenario serves as a reference point for evaluating the impact of robot assistance.

\textbf{2) Reactive Robot Assistance:} Under this policy, the robot intervenes when a human worker makes a mistake. However, the robot's intervention is not based on the human worker's expertise or its influence over the worker.
This policy, while simple, represents a common approach where the robot intervenes in response to human errors. This baseline thus provides a clear contrast to our expertise-influence-based policy.


\subsection{Results}
\begin{figure}
\centering
\includegraphics[width=0.95\columnwidth]{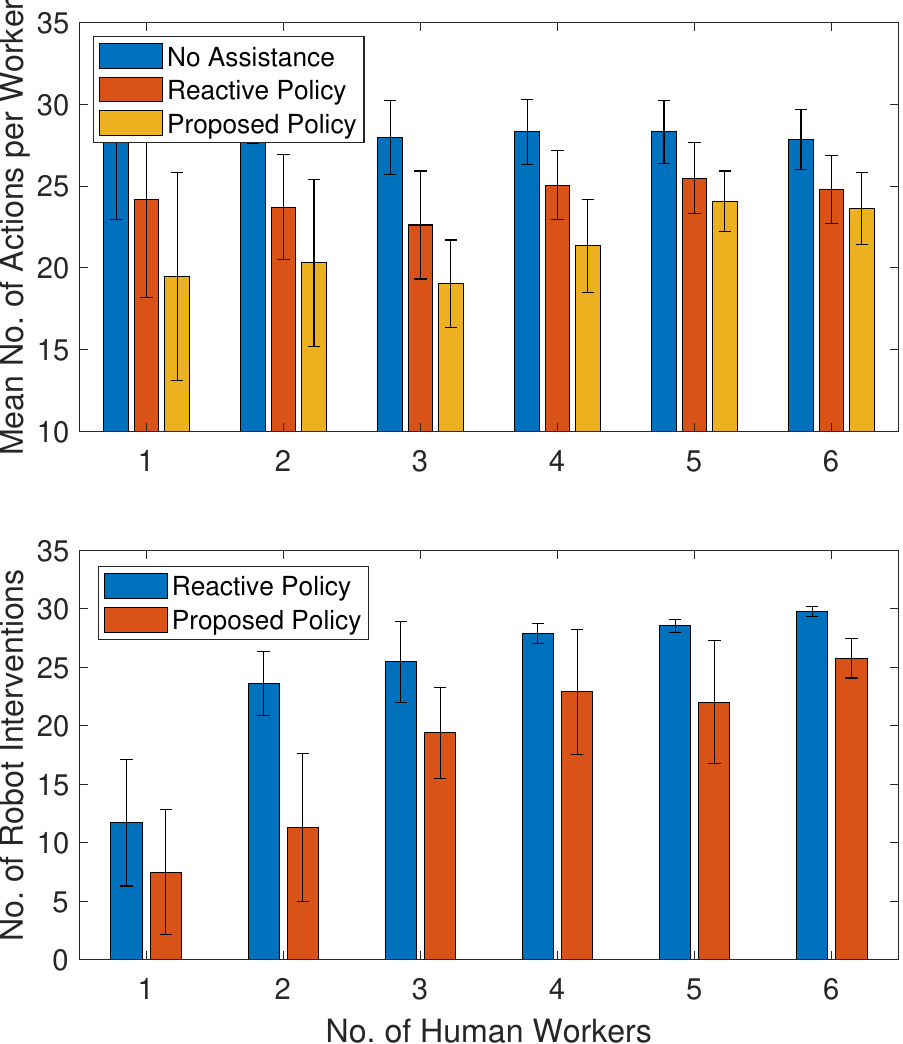}
\caption{Performance variation with increasing numbers of human workers in the kit fulfillment task. The top figure shows the mean number of actions taken by human workers to complete the task, while the bottom figure shows the mean number of robot interventions (suggested actions). The error bars show one standard deviation above and below the mean. Data shows the mean of all $30$ test runs with different sets of human workers sampled randomly from the pool.\vspace{-1em}}
\label{fig:random_sample}
\end{figure}

\begin{figure}
\centering
\includegraphics[width=0.9\columnwidth]{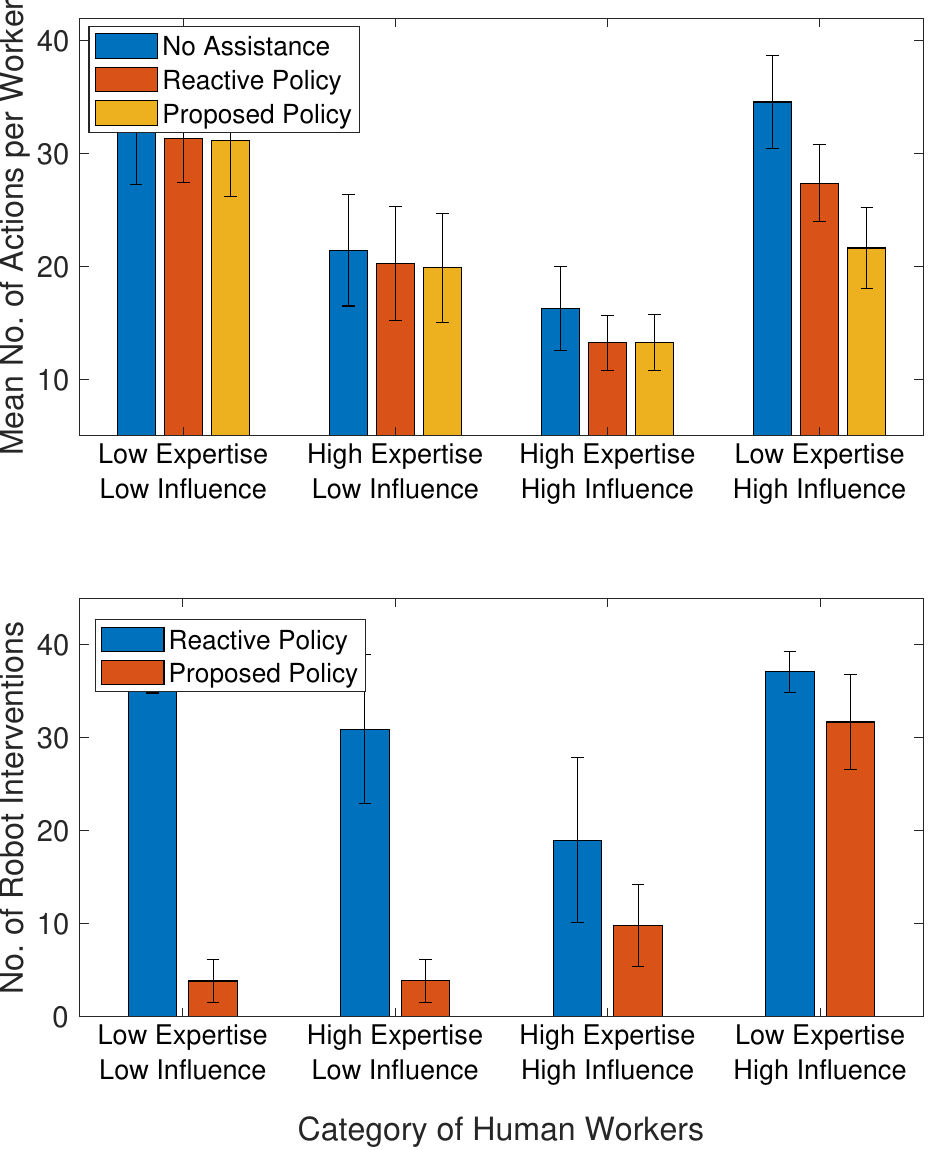}
\caption{Performance and robot assistance variations with different categories of human teammates. The top figure displays the mean number of human actions and the bottom figure shows the mean number of robot interventions. Human teammates are categorized into groups with low expertise ($\beta \in [0.1,0.5]$), high expertise ($\beta \in [2.0,2.5]$), low influence ($\theta \in [0.5,0.55]$), and high influence ($\theta \in [0.8,0.9]$).\vspace{-1em}}
\label{fig:high_low}
\end{figure}

Figures~\ref{fig:random_sample} and \ref{fig:high_low} present a performance comparison of the three policies applied to the kit fulfillment task. The task instance chosen can be optimally completed in $10$ human actions. To conduct these simulations, we established a pool of $12$ human workers, each with behavioral parameters sampled as detailed in Table~\ref{table_values}. For each test instance, we varied the number of human workers from $1$ to $6$, randomly selecting them from the pool. We ran simulations for $5$ trials under all three policies for each configuration. This process was repeated $6$ times, with each iteration involving a new set of workers, resulting in a total of $30$ runs. The presented graphs display the mean and standard deviation computed from these $30$ runs.

Figure~\ref{fig:random_sample} illustrates how the three policies performed with randomly sampled human teammates. The proposed Attention-Switching policy, on average, resulted in a $24.48\%$ reduction in human actions compared to the no-assistance case and a $12.35\%$ reduction compared to the reactive policy. Additionally, our policy led to a $25.94\%$ reduction in interventions compared to the reactive policy.

In Figure~\ref{fig:high_low}, we analyze the robot's interactions with specific categories of human teammates, considering expertise and influence parameters at the extremes of the population. The reactive assistance policy tends to intervene regardless of human behavioral parameters, whereas our policy selectively assists individuals who can benefit the most from robot assistance. With our policy, humans with low influence received an average of only $3.86$ interventions, while humans with high influence and low expertise received the most assistance, averaging $31.685$ interventions during test runs.

It's worth noting that while the proposed policy improved overall team performance, its relative benefits diminished with an increasing number of humans. This outcome was expected since the robot can only assist one person at a time, and a increasing number of humans resulted in a diminishing impact on the team's performance.


\section{Conclusions and Future work/Discussion}
\label{sec:conclusions}
This paper introduces a formal framework for robot assistance planning within the context of task-planning problems featuring multiple human teammates. The effectiveness of this framework is demonstrated through a simulated kit fulfillment task. This formal representation expands upon existing human-robot interaction frameworks by accommodating human behavioral parameters and scenarios with multiple humans, thus empowering robotic systems to provide simultaneous assistance to multiple individuals.

Future directions for this research involve exploring alternative interaction models that can enhance the prediction of human behavioral parameters, thereby improving the quality of robot assistance. Additionally, a user study is planned to validate the expertise-influence model by investigating its ability to capture real human-robot interactions accurately. Furthermore, the incorporation of observed interactions with neighboring humans to model behavioral dynamics presents an intriguing avenue for future exploration in this domain.




\end{document}